\documentclass{bmvc2k}

\usepackage{multirow}
\usepackage{caption}
\usepackage{mathtools}
\usepackage[utf8]{inputenc}

\usepackage{float}
\usepackage{amsfonts}
\usepackage{multicol}
\usepackage{booktabs}
\title{Initial Classifier Weights Replay for Memoryless Class Incremental Learning}

\addauthor{Eden Belouadah}{eden.belouadah@cea.fr}{12}
\addauthor{Adrian Popescu}{adrian.popescu@cea.fr}{1}
\addauthor{Ioannis Kanellos}{ioannis.kanellos@imt-atlantique.fr}{2}

\addinstitution{
 CEA, LIST,\\
 F-91191, Gif-sur-Yvette, France
}
\addinstitution{
 IMT Atlantique,\\
 Computer Science Department,\\
 F-29238, Brest, France
}

\runninghead{Belouadah, Popescu, Kanellos}{Initial Classfier Weights Replay...}

\begin{document}

\maketitle

\begin{abstract}
Incremental Learning (IL) is useful when artificial systems need to deal with streams of data and do not have access to all data at all times.
The most challenging setting requires a constant complexity of the deep model and an incremental model update without access to a bounded memory of past data.
Then, the representations of past classes are strongly affected by catastrophic forgetting.
To mitigate its negative effect, an adapted fine tuning which includes knowledge distillation is usually deployed.
We propose a different approach based on a vanilla fine tuning backbone.
It leverages initial classifier weights which provide a strong representation of past classes because they are trained with all class data.
However, the magnitude of classifiers learned in different states varies and normalization is needed for a fair handling of all classes.
Normalization is performed by standardizing the initial classifier weights, which are assumed to be normally distributed.
In addition, a calibration of prediction scores is done by using state level statistics to further improve classification fairness.
We conduct a thorough evaluation with four public datasets in a memoryless incremental learning setting. 
Results show that our method outperforms existing techniques by a large margin for large-scale datasets. 

\end{abstract}

%-------------------------------------------------------------------------
\section{Introduction}
\label{sec:intro}
Artificial agents are often deployed in applications which need to work under strong computational constraints and receive data sequentially.
Incremental Learning (IL) algorithms are deployed to deal with such situations. Examples include (1) exploring robots that receive data in streams and that have a limited access to a memory or no memory, (2) disease classification systems that are not allowed to access past data for privacy issues and, (3) tweets analysis where new data arrives at a fast pace and should be handled in a timely manner.
Recent approaches to class IL~\cite{DBLP:conf/eccv/AljundiBERT18,DBLP:conf/eccv/LiH16,DBLP:conf/cvpr/RebuffiKSL17,DBLP:conf/cvpr/HouPLWL19,DBLP:conf/cvpr/WuCWYLGF19} are built using deep neural networks at their core. 
The main challenge faced in IL is catastrophic forgetting~\cite{mccloskey:catastrophic}, i.e., the tendency of neural networks to forget previously learned information upon ingesting new data. The most important three characteristics that qualify an IL system to be effective and efficient are the low dependency on memory, the high accuracy on both past and new data, and the time needed to update the model to incorporate new data. 

Class IL is even more challenging when deep model complexity needs to stay constant over time and access to past data is impossible. 
Most existing works relax either the complexity requirement to allow supplementary parameters in the model~\cite{DBLP:conf/eccv/AljundiBERT18,DBLP:journals/corr/RusuRDSKKPH16,DBLP:conf/cvpr/WangRH17,DBLP:conf/eccv/MallyaDL18} or the memory constraint to store a limited number of samples of past classes~\cite{DBLP:conf/cvpr/RebuffiKSL17,DBLP:conf/cvpr/HouPLWL19,DBLP:conf/cvpr/WuCWYLGF19,javed2018,DBLP:conf/bmvc/He0SC18,scail2020}.
The setting in which deep model complexity is constant was first studied in $iCaRL$~\cite{DBLP:conf/cvpr/RebuffiKSL17}.
The authors notably adapted \textit{Learning without Forgetting} ($LwF$)~\cite{DBLP:conf/eccv/LiH16} to an IL setting to mitigate catastrophic forgetting via the use of knowledge distillation~\cite{DBLP:journals/corr/HintonVD15}.
One influential finding from~\cite{DBLP:conf/cvpr/RebuffiKSL17} was that vanilla fine tuning ($FT$) is unfit for IL but this finding was recently challenged for IL with bounded memory of the past~\cite{ahn2020simple,scail2020}.

\begin{figure}[t]
	\begin{center}
\includegraphics[width=0.99\textwidth,trim={1cm 10cm 0cm 0.1cm}]{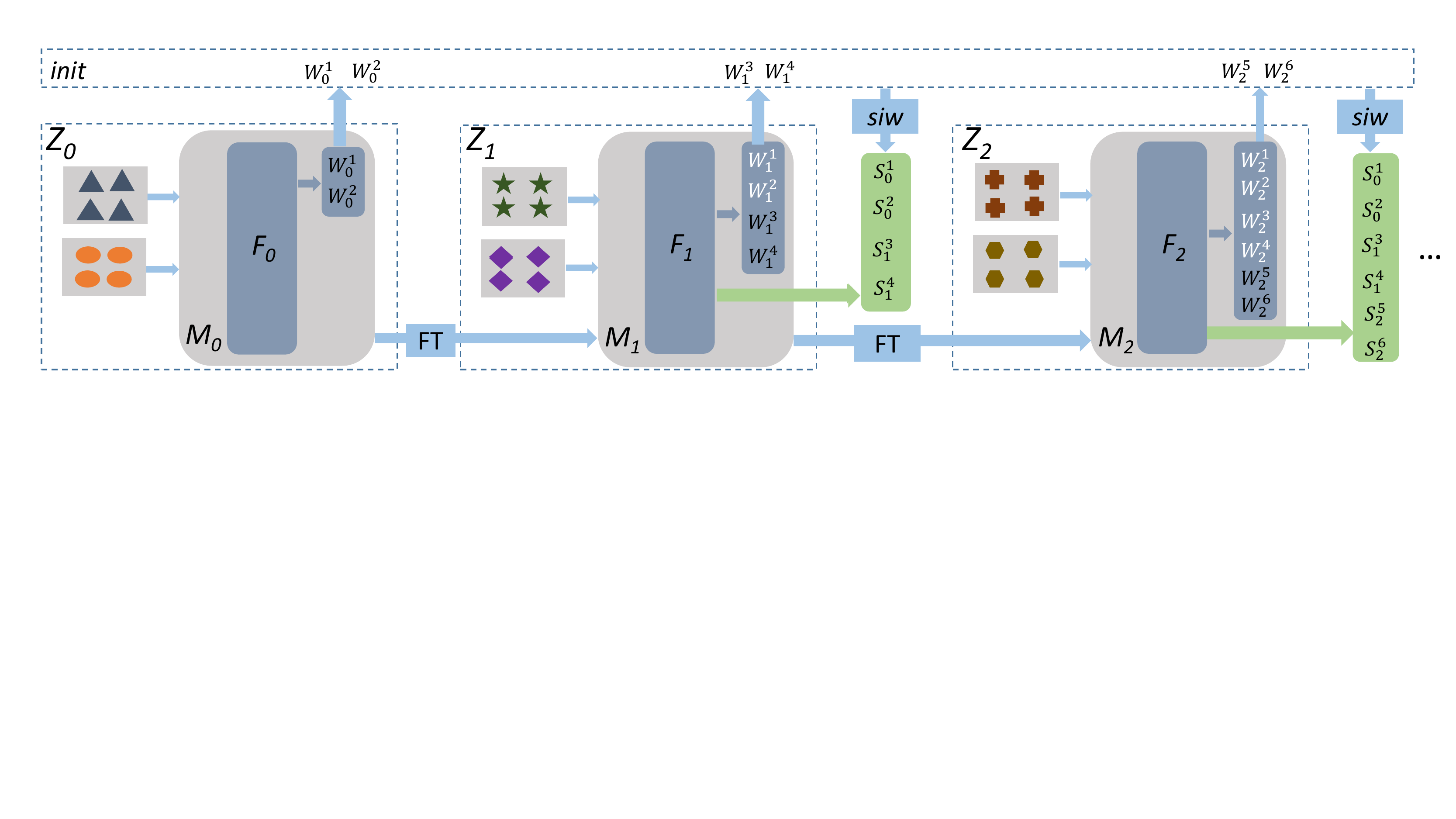}
	\end{center}
	\caption{Overview of the proposed method with an initial and two incremental states, and two classes per state. 
	A deep model $\mathcal{M}_t$ is trained for each state $\mathcal{Z}_t$. 
	Each model includes a feature extractor $\mathcal{F}_t$ and a classification layer $\mathcal{W}_t$. In memoryless IL, models have access only to data for new classes and to the previous model for fine tuning ($FT$).
	Note that $FT$ can be implemented in different ways: with a classical distillation component to counter catastrophic forgetting~\cite{DBLP:conf/cvpr/RebuffiKSL17,DBLP:conf/eccv/CastroMGSA18,javed2018}, with a more sophisticated tackling of forgetting~\cite{DBLP:conf/cvpr/HouPLWL19} or using vanilla $FT$~\cite{belouadah2019il2m,scail2020}. 
	Existing end-to-end methods~\cite{DBLP:conf/eccv/CastroMGSA18,DBLP:conf/cvpr/HouPLWL19} (in blue) perform recognition using the weights from $\mathcal{W}_t$, the classification layer of the current model which updates past classifiers. 
	Our approach (in green) is different because it freezes classifiers learned initially ($init$) for each class and applies standardization ($siw$) to make them more comparable.
	}
	\label{fig:overview}
\end{figure}

Memoryless class IL is understudied in litterature but important in practice when past data are impossible to store (due to confidentiality issues for example). We tackle memoryless class IL and provide overview of the introduced method in Figure~\ref{fig:overview}.
Our main hypothesis is that catastrophic forgetting mainly affects the classification layer of deep models during vanilla fine tuning.
Consequently, we propose to exploit initial classifier weights learned with all data of past classes.
Initial weights provide a good representation of past classes but need normalization for use in later IL states because their magnitude varies significantly across IL states.  
Preliminary analysis indicates that the magnitude of classifiers tends to decrease for classes which are learned in later states.
We exploit classifier standardization as a way to normalize initial classifiers.
%We ensure fairness for classifiers learned in different states by normalizing them with methods such as L2-normalization and variable standardization.
Normalization helps but the prediction scores also change due to variable performance of IL models.
Consequently, we also introduce a state-level calibration of class predictions inspired from~\cite{belouadah2019il2m}.
We evaluate results with four public datasets and three values for the number of incremental states.
Results show that our approach performs consequently better than competitive baselines for large scale datasets.

\section{Related Work}
\label{sec:sota}
Incremental learning is a longstanding topic in machine learning~\cite{cauwenbergs01incrementaldecremental,ruping2001} and has recently regained traction with the introduction of deep learning-based methods~\cite{DBLP:conf/cvpr/AljundiCT17,DBLP:conf/cvpr/RebuffiKSL17,DBLP:conf/eccv/CastroMGSA18}.
The main challenge is to counter the negative effects of catastrophic forgetting~\cite{mccloskey:catastrophic} by finding a good compromise between stability and plasticity of the learned models.
IL can be modeled so as to sequentially integrate new tasks~\cite{DBLP:journals/corr/RusuRDSKKPH16,DBLP:conf/cvpr/AljundiCT17} or to deal with mixed streamed data without task boundaries~\cite{DBLP:conf/cvpr/RebuffiKSL17,DBLP:conf/cvpr/HouPLWL19}. 
The first scenario assumes that new data all belong to a coherent task while the second makes the more natural hypothesis that new data are distributed across different tasks~\cite{Oswald2020Continual}.
Our work is more adapted to the second scenario since no prior is used concerning the order of classes or the boundaries between tasks.

One set of approaches addresses the problem by augmenting deep model complexity to incorporate new tasks. 
An early approach was introduced in~\cite{DBLP:journals/corr/RusuRDSKKPH16}.
It copies the latest model available and adds a part that encodes information for the new task.
The method is effective but requires a relatively large number of parameters dedicated to each task.
$PackNet$~\cite{DBLP:conf/cvpr/MallyaL18} and $Piggyback$~\cite{DBLP:conf/eccv/MallyaDL18} require less additional parameters by exploiting network pruning techniques to identify redundant parameters and reassign them to new tasks. 
An important limitation of $PackNet$ and $Piggyback$ is that they only work with small sequences of tasks. 
A hypernetwork based approach was proposed very recently to optimize the number of parameters which are stored in memory for each task~\cite{Oswald2020Continual}.
While interesting, these approaches have a scalability issue because each incremental step augments the complexity of the deep model. 

Closer to our approach are methods which keep the size of the model constant during IL and replays exemplars from a bounded memory to mitigate catastrophic forgetting.
These methods have a long history~\cite{Robins95catastrophicforgetting} and were revived in a deep learning context starting with $iCaRL$~\cite{DBLP:conf/cvpr/RebuffiKSL17}.
This method addresses three core components of replay-based IL methods: (1) the mechanism which offers a trade-off between model stability and plasticity, (2) the method used to store representative exemplars in memory and (3) the classification layer which should reduce the prediction bias toward new classes.
A knowledge distillation loss term~\cite{DBLP:journals/corr/HintonVD15,DBLP:conf/eccv/LiH16} is used to stabilize the representation of past classes while also incorporating new classes.
A nearest-class-mean
classifier~\cite{DBLP:journals/pami/MensinkVPC13} is adapted to reduce this bias.
Inspired by $iCaRL$, many subsequent replay-based IL methods focused on improving one or several of the three components. 
The authors of~\cite{javed2018} change the knowledge distillation from $iCaRL$ to a form that is closer to the original one from~\cite{DBLP:journals/corr/HintonVD15} and report an improved overall performance of the method.
More elaborate mechanisms were proposed recently to improve the trade-off between stability and plasticity.
%$M2KD$~\cite{DBLP:journals/corr/abs-1904-01769} distills knowledge both from the latest state and from the initial state of the classes.
%A second term is introduced to distill knowledge on the intermediate layers of the network.  
In~\cite{DBLP:journals/access/XiangMCX20}, the authors deploy an algorithm to compute a dynamic vector which corrects the bias induced by distillation loss among past classes and improves the representativeness of past image features.
$LUCIR$~\cite{DBLP:conf/cvpr/HouPLWL19} is a recent method that started gaining traction in the community.
It combines three components to ensure fairness between past and new classes: (1) cosine normalization acts on the magnitudes of past and new class predictions, (2) less-forget constraint modifies the usual distillation loss to handle feature vectors instead of predictions and (3) inter-class separation induces the creation of a large margin between past and new classes.

The importance of the classification layer was emphasized in~\cite{DBLP:conf/eccv/CastroMGSA18}, where a balanced fine-tuning step was introduced to reduce the bias toward new classes.
An elegant solution which adds a linear layer for bias removal was proposed in~\cite{DBLP:conf/cvpr/WuCWYLGF19}.
This method depends on a validation set and is very effective for relatively large memory sizes.
However, its performance drops sharply when the size of the validation set becomes insufficient~\cite{scail2020}.
$ScaIL$~\cite{scail2020} is more related to our work since it reduces bias by reusing the classifier weights learned initially with all data.
However, the method is unusable in memoryless IL since it relies on the comparison of classifier weights from the initial and the current state.
We note that $ScaIL$ uses vanilla fine tuning as backbone for IL and reported results are competitive with recent methods that exploit knowledge distillation~\cite{DBLP:conf/cvpr/HouPLWL19,DBLP:conf/cvpr/WuCWYLGF19}.
A similar finding was very recently reported in~\cite{ahn2020simple}, where the knowledge distillation term is also ablated.

We build on existing works to: exploit information from the initial state of classifiers~\cite{DBLP:journals/corr/abs-1904-01769,scail2020}; ensure fairness for the predictions associated to past and new classes~\cite{DBLP:conf/cvpr/HouPLWL19,DBLP:conf/cvpr/WuCWYLGF19,belouadah2019il2m}; use vanilla $FT$ as a backbone to train deep models~\cite{ahn2020simple,scail2020} .
However, our approach differs through the method used for the normalization of initial classifier weights and the focus is on a large scale memoryless IL.
Note that many existing methods cannot operate in the absence of memory and become unusable in our setting.
The following ones are usable in memoryless IL and will be used in the evaluation: $LwF$~\cite{DBLP:conf/eccv/LiH16}, the end-to-end version of $LUCIR$~\cite{DBLP:conf/cvpr/HouPLWL19}, baseline methods which exploit initial classifiers without bounded memory from~\cite{scail2020}.

\section{Proposed Method}

\subsection{Motivation}
\label{subsec:analysis}
We hypothesize that it is possible to exploit initial classifier weights, learned when data is first streamed for each class, to mitigate catastrophic forgetting in memoryless IL.
CNN prediction scores are obtained from the combination of features provided by the penultimate layer of the model with classifier weights from the final layer.
We analyze these two layers in an IL context to motivate our approach. 
The ILSVRC dataset with an initial and nine incremental states is used for both analyses.

\begin{figure}[h!]
\centering
\begin{minipage}{.5\textwidth}
  \centering 
  \captionsetup{width=.9\linewidth}
  \includegraphics[width=.999\linewidth]{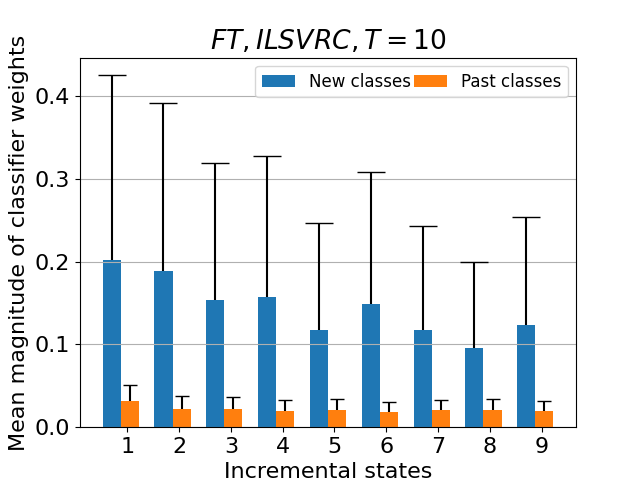}
  
  \vspace{0.4em}
  
  \captionof*{figure}{(a)}
  \label{fig:gauche}
\end{minipage}%
\begin{minipage}{.5\textwidth}
  \centering
    \captionsetup{width=.9\linewidth}
  \includegraphics[width=.999\linewidth]{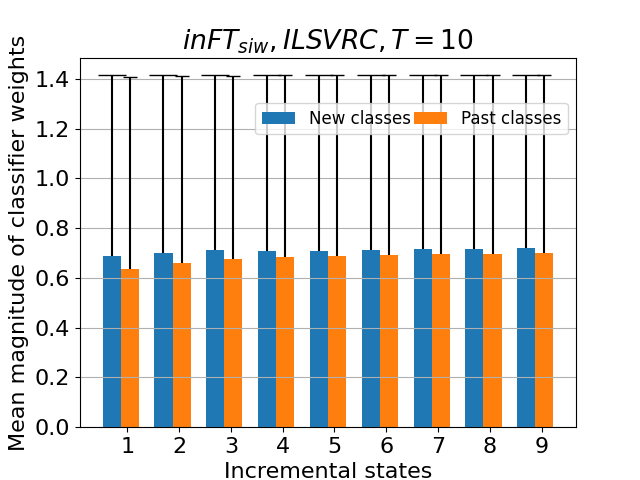}
  
  \vspace{0.4em}
  
  \captionof*{figure}{(b)}
  \label{fig:droite}
\end{minipage}

\vspace{1em}

\caption{Mean magnitudes of classifier weights for new and past classes in memoryless incremental learning: (a) left - with vanilla fine tuning affected by catastrophic forgetting and (b) right -  after standardization of initial classifiers. Mean magnitudes are computed only for incremental states and the first non-incremental state is excluded. Note that the reference state is the rightmost one in each figure.
}
\label{fig:weights}
\end{figure}

In Figure~\ref{fig:weights}(a), we present a summarized view of the magnitudes of classifiers for an IL baseline which exploits vanilla fine tuning.
The absolute values of individual classifier dimensions are aggregated to compute mean magnitudes for new and past classes respectively.
Past classes have much lower magnitudes because there are no exemplars to be replayed for them in memoryless IL.
Since magnitudes are much higher for new classes, test examples will always be attributed to one of these classes even if they belong to past classes.
This observation provides further support for the previous conclusion that vanilla $FT$ is not directly usable in IL~\cite{DBLP:conf/cvpr/RebuffiKSL17,DBLP:conf/eccv/CastroMGSA18}.

The magnitudes of new classes in Figure~\ref{fig:weights}(a) vary across incremental states, with a global tendency toward reduction in later states.
A normalization of initial classifiers is thus needed to ensure fairness if they are replayed across incremental states as proposed here. 
New classifiers from previous states of Figure~\ref{fig:weights}(a) are aggregated to represent past classes in each current state of Figure~\ref{fig:weights}(b).
Normalization makes statistical populations more comparable and it is obtained by applying standardization~\cite{stats2007}, a method which is discussed in detail in Subsection~\ref{subsec:cwadapt}.
The standardized classifiers, illustrated in Figure~\ref{fig:weights}(b), have comparable magnitudes and become usable in memoryless IL.

\begin{figure}[h]
\minipage{0.245\textwidth}
  \includegraphics[width=\linewidth]{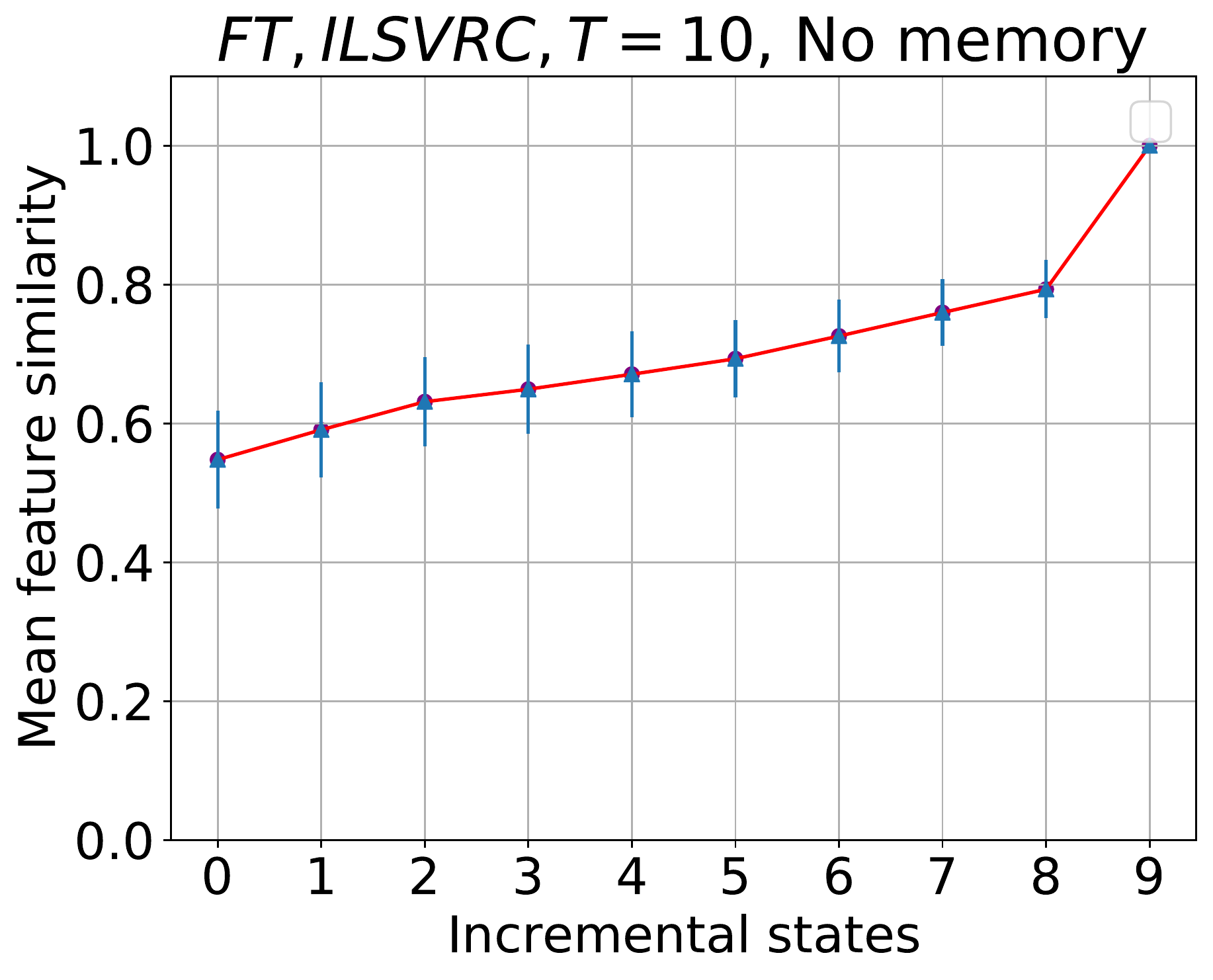}
  %\caption{A really Awesome Image}\label{fig:awesome_image1}
    \vspace{0.4em}
  
  \captionof*{figure}{(a)}
\endminipage\hfill
\minipage{0.245\textwidth}
  \includegraphics[width=\linewidth]{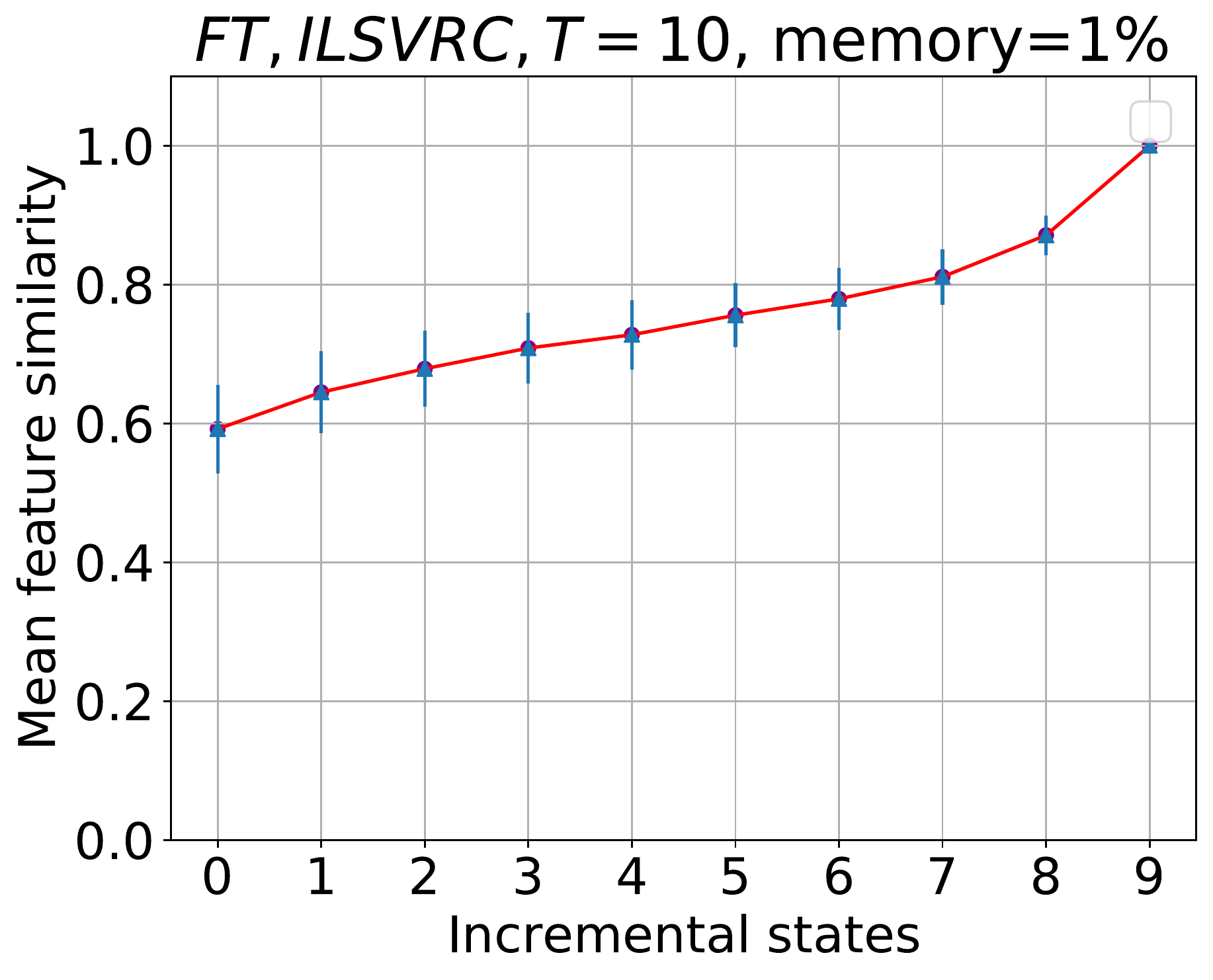}
  %\caption{A really Awesome Image}\label{fig:awesome_image2}
    \vspace{0.4em}
  
  \captionof*{figure}{(b)}
\endminipage\hfill
\minipage{0.245\textwidth}%
  \includegraphics[width=\linewidth]{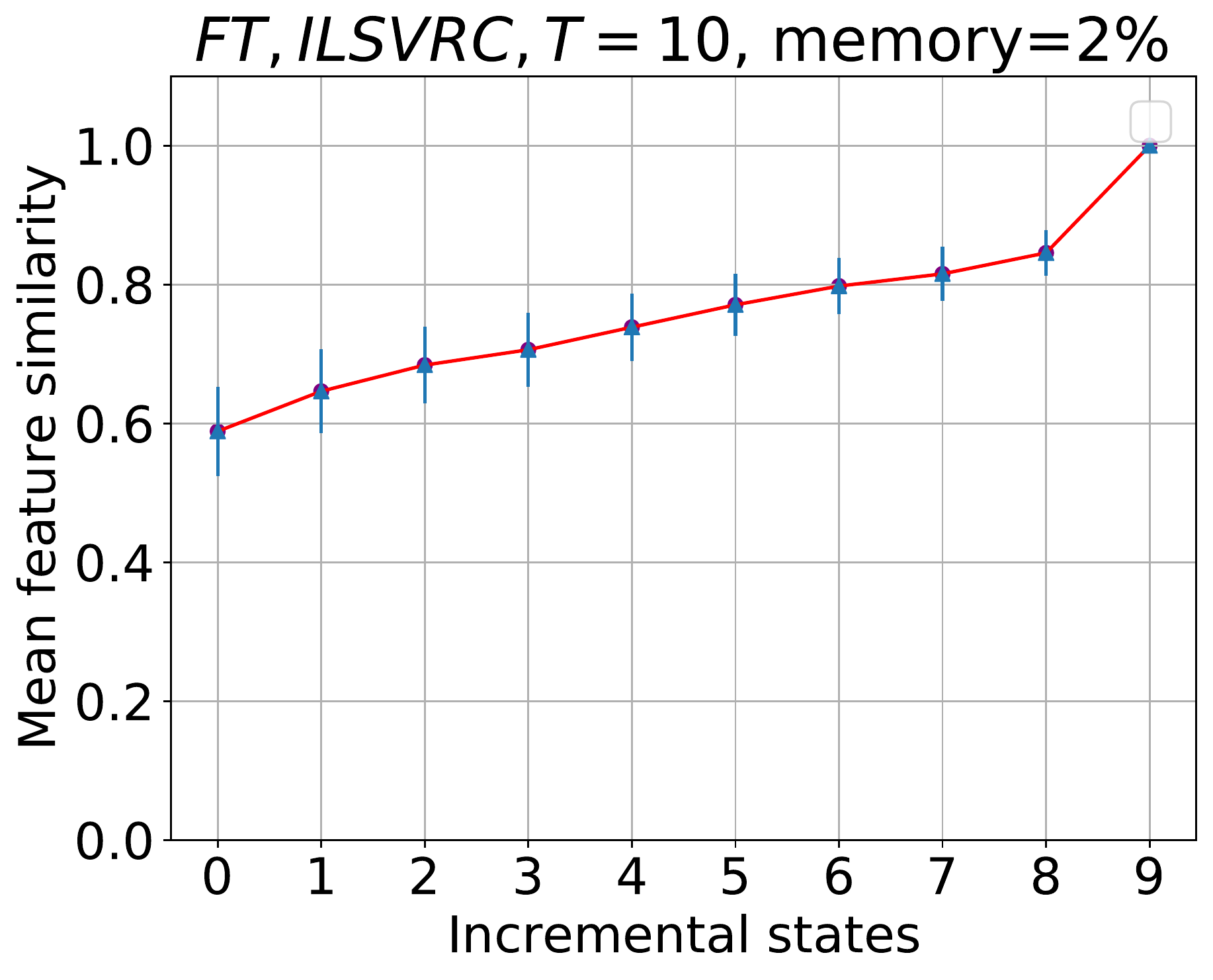}
  %\caption{A really Awesome Image}\label{fig:awesome_image3}
    \vspace{0.4em}
  
  \captionof*{figure}{(c)}
\endminipage
\minipage{0.245\textwidth}%
  \includegraphics[width=\linewidth]{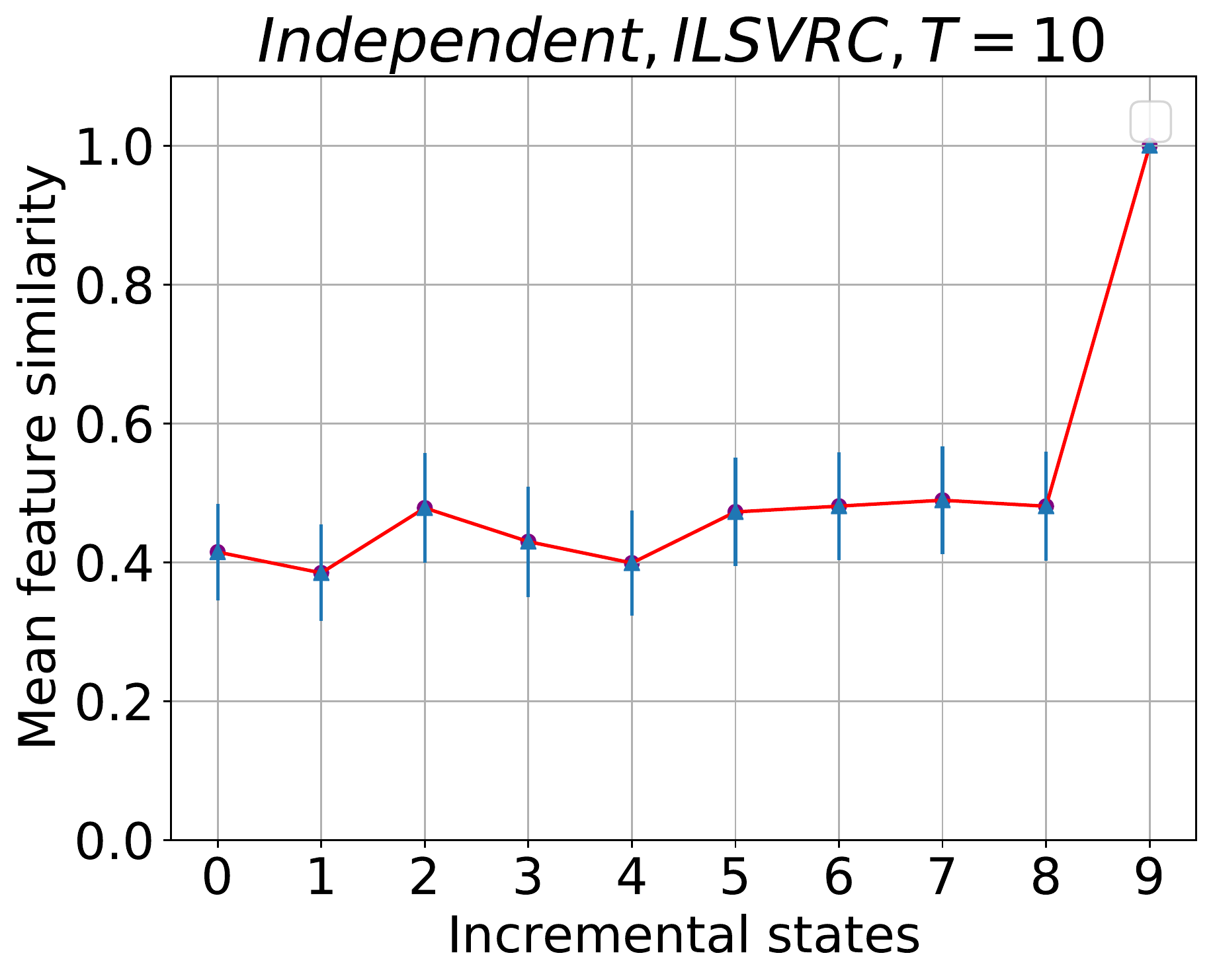}
  %\caption{A really Awesome Image}\label{fig:awesome_image3}
    \vspace{0.4em}
  
  \captionof*{figure}{(d)}
\endminipage

\vspace{1em}

\caption{Mean feature similarities between incremental states for test images of the first state. 
Cosine similarities are computed for vanilla fine tuning as follows: (a) without memory, (b) with bounded memory of 1\% of the dataset, (c) with bounded memory of 2\%. 
Subfigure (d) plots lower-bound similarities for the case when individual states are learned independently, without fine tuning.
The upper bound for similarity is 1 and would be obtained for the model which is frozen after the initial state.
However, such an approach has no plasticity and cannot incorporate knowledge related to data streamed during IL. 
The final incremental state (9) is used as reference to compute similarities with other states.
The more distant two states are, the lower the similarity is likely to be. \\
}
\label{fig:features}
\end{figure}

A second important assumption of our approach is that features extracted from the penultimate layer of the current IL model are compatible with initial classifiers from previous states.
This assumption holds if the current features keep a trace of what was learned before.
We design a simple experiment that assesses the degree of similarity between features of the same images extracted in different incremental states.
Features are extracted for test images of the initial non-incremental state using models learned in each incremental state.
Features of test images extracted in the $9^{th}$ and last incremental state are used as reference to illustrate the use of initial classifiers from previous states with its features.
Cosine similarity between them and the features of the same images from each previous state is computed.
The mean feature similarities between the last state and previous ones are presented in Figure~\ref{fig:features}.

To better situate similarities for memoryless IL (Figure~\ref{fig:features}(a)), we also present statistics for IL with bounded memory, including 1\% and 2\% of the dataset (Figure~\ref{fig:features}(b) and (c) respectively).
We also provide similarities for independent training of incremental states where no fine tuning is used in Figure~\ref{fig:features}(d).
Naturally, this last setting is not a valid IL approach and is shown only as an illustration for a lower bound of feature similarity.

The results from Figure~\ref{fig:features} indicate that mean similarities obtained with fine tuning without and with memory are significantly higher than those obtained with independent training.
This finding validates the fact that current features learned with vanilla $FT$ keep a trace of what was learned in previous states.
Mean similarities decrease with the distance between the current state and the initial one because forgetting is higher when more trainings are involved.
The use of a bounded memory in Figure~\ref{fig:features}(b) and (c) provides better similarities compared to memoryless IL in Figure~\ref{fig:features}(a).
The effect is particularly visible for states such as 8 or 7 which are close to the reference one and becomes less important for more distant states such as 2 or 1.
Note that features from the current IL state were used successfully with initial classifiers in~\cite{scail2020} if a bounded memory of the past was allowed.
The comparison of feature similarities indicates that their use with adapted initial classifiers is more challenging without memory than with a bounded memory but still doable.

\subsection{Memoryless IL with Normalized Initial Classifier Weights}
\label{subsec:cwadapt}
We build on previous works~\cite{cauwenbergs01incrementaldecremental,DBLP:conf/cvpr/RebuffiKSL17,DBLP:conf/eccv/CastroMGSA18, scail2020} to define memoryless class incremental learning.
We note $T$ the total number of states, including the first non-incremental state and $T-1$ incremental states.  
Data arrive in a stream which includes $n_0 = n_1 = n_2 = ...  = n_t$ classes for each state of the class incremental process.
A first model $\mathcal{M}_0$ is trained from scratch on a set of $n_0$ classes in the initial non-incremental state $\mathcal{Z}_0$. 
The incremental models $\mathcal{M}_1,...,\mathcal{M}_t,...,\mathcal{M}_T$ are trained in states $\mathcal{Z}_1,..., \mathcal{Z}_t,...,\mathcal{Z}_T$. 
Each incremental model $\mathcal{M}_t$ ($t=1 ... T$) is initialized using the weights of $\mathcal{M}_{t-1}$ and accesses training data streamed for the current $n_t$ new classes.
Since no memory of the past is allowed, no data are available for past classes but $\mathcal{M}_t$ should be able to recognize all classes seen so far $N_t=n_0+n_1+...+n_t$.
Models include a feature extractor $\mathcal{F}_t$ which produces $d$-dimensional features $f_t$ for each image and a classification layer.
A standard CNN model learned in the $t^{th}$ state of an IL process recognizes image content using a classification layer made of a classifier weights matrix $\mathcal{W}_t$ and a corresponding bias vector $\mathcal{B}_t$.
$\mathcal{B}_t$ is less important than $\mathcal{W}_t$ and we focus the discussion on the weights matrix. $\mathcal{W}_t$ is defined as:
\begin{equation}
\mathcal{W}_t = \{W_t^{1},..,W_t^{n_{0}}, 
W_t^{n_{0}+1},..,W_t^{n_{1}},
...,
W_{t}^{n_{t-2}+1},..,W_{t}^{n_{t-1}},
W_t^{n_{t-1}+1},..,W_t^{n_{t}}
\}
\label{eq:weights}
\end{equation}
 
We propose to replay the initial classifiers of each class in order to mitigate catastrophic forgetting.
The classification layer made of initial classifier weights is written as:
\begin{equation}
\mathcal{W}_t^{in} = \{W_0^{1},..,W_0^{n_{0}}, 
W_1^{n_{0}+1},..,W_1^{n_{1}},
...,
W_{t-1}^{n_{t-2}+1},..,W_{t-1}^{n_{t-1}},
W_t^{n_{t-1}+1},..,W_t^{n_{t}}
\}
\label{eq:inweights}
\end{equation}

The analysis from Subsection~\ref{subsec:analysis} shows that the weights from Eq.~\ref{eq:inweights} need normalization to become comparable across states.
We apply a standardization of initial weights ($siw$) to obtain a normalized version of the weights matrix:

\begin{equation}
\mathcal{S}_t^{in} = \{S_0^{1},..,S_0^{n_{0}}, 
S_1^{n_{0}+1},..,S_1^{n_{1}},
...,
S_{t-1}^{n_{t-2}+1},..,S_{t-1}^{n_{t-1}},
S_t^{n_{t-1}+1},..,S_t^{n_{t}}
\}
\label{eq:siwweights}
\end{equation}

Each dimension $s_{k}$ of a standardized classifier $S$ from Eq.~\ref{eq:siwweights} is calculated using:

\begin{equation}
s_{k} = \frac{w_k-\mu(W)}{\sigma(W)}
\end{equation}
with: 
$s_k$ - the $k^{th}$ dimension of $S$,
$w_k$ - the $k^{th}$ dimension of an initial classifier $W$ from Eq.~\ref{eq:inweights},
$\mu(W)$ and $\sigma(W)$ are the mean and standard deviation of $W$.

\begin{figure}[!htb]
\minipage{0.32\textwidth}
  \includegraphics[width=\linewidth]{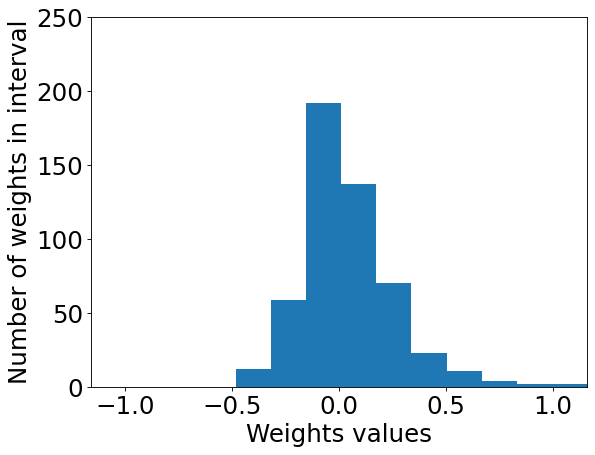}
    \vspace{0.4em}
  
\endminipage\hfill
\minipage{0.32\textwidth}
  \includegraphics[width=\linewidth]{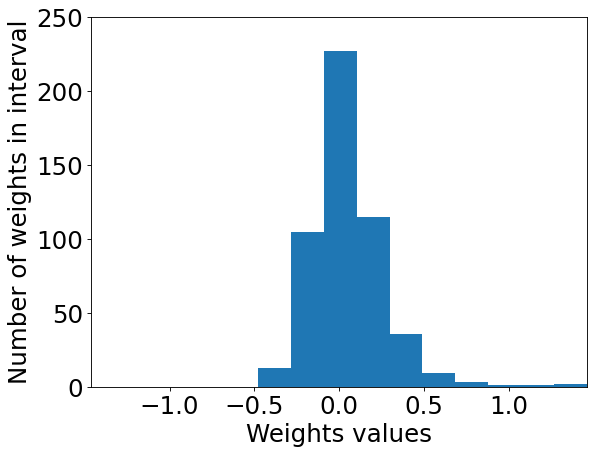}
    \vspace{0.4em}
  
\endminipage\hfill
\minipage{0.32\textwidth}%
  \includegraphics[width=\linewidth]{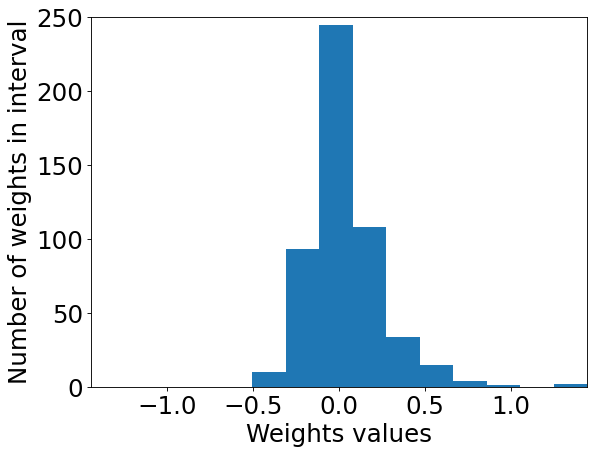}
    \vspace{0.4em}
 
\endminipage\hfill

\vspace{0.4em}

\minipage{0.32\textwidth}
  \includegraphics[width=\linewidth]{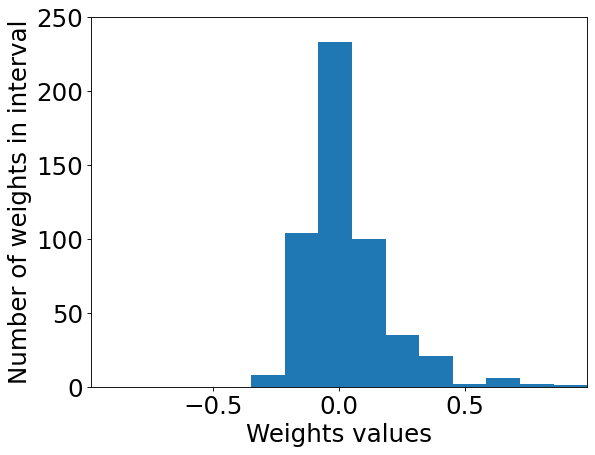}
    \vspace{0.4em}
  
\endminipage\hfill
\minipage{0.32\textwidth}
  \includegraphics[width=\linewidth]{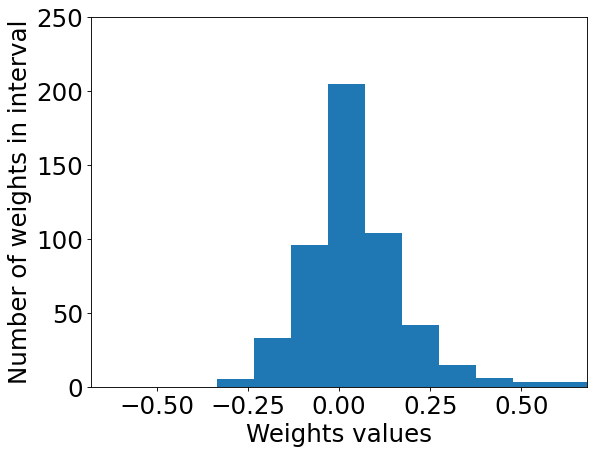}
    \vspace{0.4em}
  
  \endminipage\hfill
\minipage{0.32\textwidth}%
  \includegraphics[width=\linewidth]{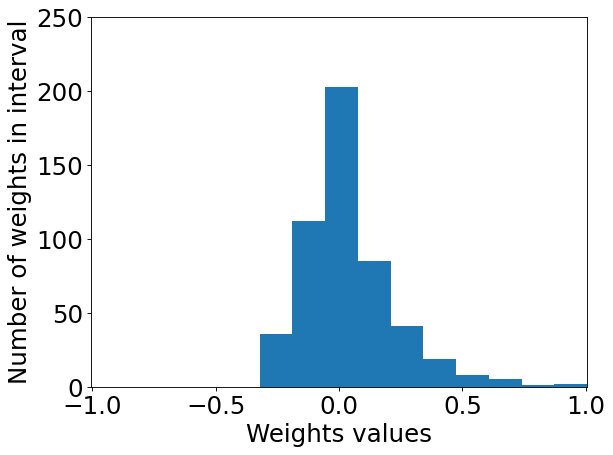}
    \vspace{0.4em}
 
\endminipage\hfill

\vspace{0.4em}

\caption{Weights distribution of a random subset of classifier weights from the weights matrix $\mathcal{W}_t$ defined in Eq.~\ref{eq:weights}. 
}
\label{fig:weights_distribution}
\end{figure}

Standardization is useful if it is applied to the statistical populations which follow a normal distribution~\cite{stats2007}, which is the case for classifier weights from Eq.~\ref{eq:inweights}.
%We present the distribution of classifier weights for a subset of classifiers in the supplementary material.
Figure~\ref{fig:weights_distribution} provides weights distribution of a random subset of classifier weights from the weights matrix $\mathcal{W}_t$ defined in Eq.~\ref{eq:weights}. 
These examples illustrate the fact that the classifier weights follow a normal distribution.
The use of standardization to normalize them is thus appropriate. 

Assuming that the incremental process is in the $t^{th}$ state, the final prediction score of a test image $x$ for $C_j^i$, the $i^{th}$ class learned initially in the $j^{th}$ state (with $j \leq t$), is given by:

\begin{equation}
    p(x,C_j^i) = (f_t(x) \cdot S_j^i + b_j^i ) \times \frac{\mu(\mathcal{M}_t)}{\mu(\mathcal{M}_j)}
    \label{eq:prediction}
\end{equation}

with:
$f_t(x)$ - features of image $x$ given by the extractor of the current model $\mathcal{M}_t$; $S_j^i$ - standardized classifier weights of the $i^{th}$ class initially learned in the $j^{th}$ state as given by Eq.~\ref{eq:siwweights}; $b_j^i$ - the class bias value; $\mu(\mathcal{M}_t)$ and $\mu(\mathcal{M}_j)$ - means of top-1 predictions of models learned in the $t^{th}$ and $j^{th}$ states computed over their training sets.

The first term of Eq.~\ref{eq:prediction} is a version of the usual CNN prediction process in which the basic weights $W_t^i$ from Eq.~\ref{eq:weights} are replaced by the standardized initial weights $S_j^i$ from Eq.~\ref{eq:siwweights}.
This term is referred to as $siw$ in Section~\ref{sec:expe}.
The second term is inspired by~\cite{belouadah2019il2m}, where the authors observed that a model level calibration is useful when combining information from different models.
$\mu(\mathcal{M}_t)$ and $\mu(\mathcal{M}_j)$ are calculated by passing all training images available in each of the two states through the respective model. 
This term is referred to as $mc$ in Section~\ref{sec:expe}.

\section{Experiments}
\label{sec:expe}

\subsection{Datasets and Implementation Details}
\label{subsec:data}
Experiments are done with four public datasets designed for different classification tasks: (1) \textbf{ILSVRC}~\cite{DBLP:journals/ijcv/RussakovskyDSKS15} - fine-grained object recognition dataset containing 1000 classes, (2) \textbf{VGGFace2}~\cite{DBLP:conf/fgr/CaoSXPZ18} - face recognition dataset containing 1000 identities, (3) \textbf{Google Landmarks}~\cite{DBLP:conf/iccv/NohASWH17} (Landmarks below) - tourist landmark recognition dataset containing 1000 classes, 
and (4) \textbf{CIFAR100}~\cite{Krizhevsky09learningmultiple} - basic level object recognition with 100 classes. Please see the supplementary material for dataset details.

To test the robustness of our system, we vary the number of states $T=\{10, 20, 50\}$. Increasing the number of states leads the model to perform more rehearsal steps. The model's parameters are more often updated which worsens forgetting~\cite{belouadah2019il2m,scail2020}. 

A ResNet18~\cite{DBLP:conf/cvpr/HeZRS16} architecture is used as backbone for FT-based methods. 
%Note that the main model hyperparameters are taken from the original implementation~\cite{DBLP:conf/cvpr/HeZRS16} without specific optimization. 
$LwF$ and $LUCIR$ are run using the public implementations from~\cite{DBLP:conf/cvpr/HouPLWL19} and~\cite{DBLP:conf/cvpr/RebuffiKSL17}. 
More implementation details are provided in the supplementary material.

\subsection{Evaluated Methods and Protocol}
\label{subsec:base}
%Since not all state-of-the-art methods operate in absence on memory, we compare our method to those that are functional without it, notably:
We use the following competitive methods from literature and their adaptations as baselines:
\begin{itemize}
    \item $LwF$~\cite{DBLP:conf/eccv/LiH16} - IL method which inspired most recent approaches. Catastrophic forgetting is reduced by using a distillation loss. For comparability with our approach, we also create the following variants: $inLwF$ - $LwF$ with raw initial weights, $inLwF_{siw}$ - $inLwF$ with $siw$ (standardization of classifier weights)  from Eq.~\ref{eq:prediction} and $inLwF_{siw}^{mc}$ - $inLwF$ with standardization plus $mc$ calibration from Eq.~\ref{eq:prediction}.
    \item $LUCIR$~\cite{DBLP:conf/cvpr/HouPLWL19} - recent approach whose end-to-end version is usable for memoryless IL. $LUCIR$ uses an elaborate mechanism to reduce catastrophic forgetting. Its definition does not allow the use of initial classifier weights but we apply $mc$ to it in $LUCIR^{mc}$ for comparability. 
    \item{$inFT$}~\cite{scail2020} - uses the initial classifier weights without normalization. It was first introduced as baseline for IL with memory.
    \item $inFT_{L2}$~\cite{scail2020} - version of $inFT$ with $L2$ normalization of initial classifier weights. $inFT_{L2}^{mc}$ combines $L2$ normalization and $mc$ calibration for comparability. 
\end{itemize}

Our methods are based on $inFT$:
$inFT_{siw}$ exploits only the $siw$ term from Eq.~\ref{eq:prediction}, while $inFT_{siw}^{mc}$ exploits both the $siw$ and the $mc$ terms from Eq.~\ref{eq:prediction}. $Full$ is the performance of classical learning. This algorithm is the upper bound for all class incremental approaches.

We follow the evaluation protocol from~\cite{DBLP:conf/eccv/CastroMGSA18}, which averages accuracy over IL states (i.e., the initial non-incremental state is excluded).
Performance is measured with top-5 accuracy. 
We also use $G_{IL}$~\cite{scail2020} for a global view of performance.
$G_{IL}$ aggregates performance gaps for individual configurations which are computed as the difference between configuration accuracy and $Full$ accuracy divided by the difference between the upper bound of the accuracy measure and $Full$ accuracy.
The closer to zero $G_{IL}$ is, the better performance will be.

\subsection{Discussion of Results}
\label{subsec:results}
\begin{table*}
\begin{center}
\resizebox{\textwidth}{!}{
\begin{tabular}{|c|ccc|ccc|ccc|ccc||c|}
\hline
Dataset & \multicolumn{3}{|c|}{ILSVRC} & \multicolumn{3}{c|}{VGGFace2} & \multicolumn{3}{c|}{Landmarks}& \multicolumn{3}{c|}{CIFAR100} & \multirow{2}{*}{$G_{IL}$}  \\
\cline{1-13}
States & T=10  & T=20  & T=50  & T=10  & T=20 & T=50 & T=10  & T=20 & T=50 & T=10  & T=20 & T=50  & \\
\hline
{\small $LwF$} & 45.3 & 37.6 & 27.1 & 53.3 & 42.6 & 30.8 & 58.8 & 49.2 & 35.2 & 79.5 &  65.3 & 39.0 & -34.72\\
{\small $inLwF$} & 47.1 & 39.9 & 32.2 & 58.1 & 50.8 & 40.5 & 55.7 & 50.2 & 39.8 & 79.4 & 67.9 & 42.8 & -31.97\\
{\small $inLwF_{siw}$} & 54.0 & 45.8 & 35.1 & 70.4 & 59.3 & 45.2 & 61.0 & 53.8 & 42.2 & \textbf{80.0} & \textbf{68.8} & \textbf{44.6}  & -28.06\\
{\small $inLwF_{siw}^{mc}$} & 40.2 & 44.7 & 33.8 & 67.5 & 56.5 & 42.0 & 54.6 & 48.0 & 37.2 & 78.6 & 67.5 & 43.8 & -30.79\\
{\small $LUCIR$} &57.6 & 39.4 & 21.9 & \textbf{91.4} & 68.2 & 32.2 & \textbf{87.8} & 63.7 & 32.3 & 57.5 & 35.3 & 21.0  & -24.75\\
{\small $LUCIR^{mc}$} & 53.7 & 30.5 & 12.7 & 82.6 & 51.0 & 21.0 & 84.1 & 44.0 & 21.6 & 45.8 & 26.8 & 23.7  & -32.18\\
{\small $FT$} &20.6 & 13.4 & 7.1 & 21.3 & 13.6 & 7.1  & 21.3 & 13.6 & 7.1 & 21.3 & 13.7 & 17.4 & -54.91 \\
{\small $inFT$} &61.0 & 44.9 & 23.8 & 90.9 & 64.4 & 33.1 & 68.8 & 49.4 & 22.2 & 55.1 & 40.8 & 19.9 & -28.99\\
{\small $inFT_{L2}$ } & 51.6 & 43.3 & 34.5 & 76.8 & 66.8 & 55.1 & 61.4 & 52.5 & 39.2  & 47.5 & 39.3 & 22.5 & -26.80\\
{\small $inFT^{mc}$} & 62.0 & 39.6 & 19.2 & 78.5 & 52.3 & 27.5 & 73.3 & 44.2 & 17.7 & 57.9 & 34.2 & 18.2 & -32.75 \\
{\small $inFT_{L2}^{mc}$} & 53.6 & 42.7 & 35.6 & 86.9 & 71.4 & 53.6 & 66.2 & 52.6 & 37.9 & 52.6 & 43.1 & 18.2 & -25.02 \\
{\small $inFT_{siw}$  } & 61.6 & 51.9 & 39.9 & 84.0 & 80.6 & 61.9 & 75.1 & 62.6 & \textbf{43.2} & 56.0 & 41.8 & 22.5  & -20.97\\
{\small $inFT_{siw}^{mc}$ } & \textbf{64.4} & \textbf{54.3} & \textbf{41.4} & 88.6 & \textbf{84.1} & \textbf{62.6} & 79.5 & \textbf{64.5} & \textbf{43.2}  & 59.7 & 44.3 & 18.4  &  \textbf{-19.38} \\
\hline
$Full$ & 
\multicolumn{3}{|c|}{92.3}&
\multicolumn{3}{c|}{99.2} & \multicolumn{3}{c|}{99.1} & \multicolumn{3}{c|}{91.2}  & 
- \\
\hline
\end{tabular}
}
\end{center}

	\caption{Top-5 average IL accuracy (\%) for the different methods  with T=\{10, 20, 50\} states. $Full$ is the performance of classical learning, with all data available. Best results are in bold.
	}
\label{tab:no_memory_results}
\end{table*}

The results from Table~\ref{tab:no_memory_results} show that $inFT_{siw}^{mc}$ provides a $G_{IL}$ improvement of over $5$ points compared to $LUCIR$~\cite{DBLP:conf/cvpr/HouPLWL19}, the best baseline. 
The gain is even higher compared to $inFT_{L2}^{mc}$, the second best baseline which extends $inFT_{L2}$~\cite{scail2020}.
This difference in favor of $inFT_{siw}^{mc}$ underlines the fact that classifier weights standardization is more appropriate than $L2$ normalization for memoryless IL. A favorable comparison to two other normalization methods is provided in the supplementary material.
Results show that the $siw$ term from Eq.~\ref{eq:prediction} is useful both for $inFT$ and $LwF$ while $mc$ is beneficial for $inFT$ but degrades results for $LwF$ and $LUCIR$. 
Standardization of $inLwF$ weights has a positive effect for all datasets compared to $LwF$, especially for $T=\{20, 50\}$ states.
Moreover, even when $mc$ works, its contribution is less important than that of $siw$.

The worst results by far are obtained with $FT$, both globally and for individual configurations.
$FT$ trains well for new classes but provides nearly random results for past classes that have no exemplars available for replay in memoryless IL.
This confirms previous findings~\cite{DBLP:conf/cvpr/RebuffiKSL17,DBLP:conf/eccv/CastroMGSA18} regarding the strong effect of catastrophic forgetting on vanilla fine tuning in memoryless IL. 

The comparison of performance for different datasets indicates that distillation is useful at small scale but has lower utility or becomes even detrimental for large datasets. The usefulness of distillation for small datasets but not for large ones is an open problem. In~\cite{scail2020, javed2018}, we note that distillation induces confusions among past classes~\cite{DBLP:journals/corr/HintonVD15}. To better understand this phenomenon, we provide an analysis of the typology of errors in the supplementary material. Note that while $FT$  runs do not perform well, normalization does help for them. 

The $inLwF_{siw}$ version of $LwF$ is the best method for CIFAR100, with a large margin compared to all methods which are not based on $LwF$. 
The use of initial weights in $inLwF$ brings a $G_{IL}$ improvement of nearly $3$ points, and the addition of standardization in $inLwF_{siw}$ brings another $4$ $G_{IL}$ points gain.
$LUCIR$ has lower performance but rather comparable to that of $inFT$ based methods for CIFAR100.
The results for the three large datasets show that 
$LwF$ has second-lowest performance, after vanilla $FT$. 
The improvements over classical $LwF$ brought by standardization are much more important for the three large datasets.
$LUCIR$'s more sophisticated scheme for countering catastrophic forgetting is clearly useful compared to classical distillation from $LwF$.
The removal of the distillation component and the use of initial classifier weights in $inFT$ gives globally better results than $LUCIR$ and $LwF$ for the three large datasets.
This behavior in a large scale setting is mainly explained by the observation made in~\cite{DBLP:conf/cvpr/HouPLWL19,scail2020} regarding the high number of confusions among past classes when distillation is used. 
The use of $siw$ and $mc$ to $inFT$ is also beneficial, especially when $T$ is big for large datasets.

The number of incremental states has an important effect on IL performance~\cite{DBLP:conf/cvpr/RebuffiKSL17,DBLP:conf/eccv/CastroMGSA18,scail2020}. 
The larger the number of states is, the more challenging the process will be. 
This is confirmed by the results with $T=\{10, 20, 50\}$ states in Table~\ref{tab:no_memory_results} for all tested methods. Actually, as more incremental states are added, IL models are more prone to forgetting due to the increasing number of intermediate model updates, which causes information loss. 
Our approach does significantly better than existing methods for the three large datasets with $T=50$. Its performance reaches $41.4$ for ILSVRC compared to only $27.1$ and $21.9$ for $LwF$ and $LUCIR$.

Table~\ref{tab:no_memory_results} allows an ablation analysis of our approach.
Such an analysis is important to understand the contribution of individual components to the final results.
Vanilla $FT$ is the backbone upon which we build. 
The use of raw initial weights for past classes in $inFT$ has a strong positive effect as it reduces the performance gap measured by $G_{IL}$ by nearly a half ($-28.99$ vs. $-54.91$ for vanilla $FT$).
The sole use of calibration in $inFT^{mc}$ has a negative effect since performance drops to $-32.75$.
The introduction of standardization in $inFT_{siw}$ has a important positive effect since it brings an $8$ points $G_{IL}$ improvement over $inFT$.
Finally, the use of both terms from Eq.~\ref{eq:prediction} in $inFT_{siw}^{mc}$ has a light positive effect with a $1.6$ points improvement over $inFT_{siw}$.
We note that $inFT_{siw}$ improves over $inFT$ for all individual configurations. 
The largest performance gains between the two methods are obtained for the three large datasets with the $T=50$, the highest number of incremental states tested.
This is the most challenging setting since the effects of catastrophic forgetting are stronger for a larger number of IL states. 
The addition of state mean calibration has a positive, albeit smaller, effect in all individual configurations but two.
It does not improve results for Landmarks with $T=50$ states and degrades them for CIFAR100 with the same number of states.
This is probably an effect of the fact that there are only two classes per state in the latter configuration and the obtained statistics are not stable enough.

\vspace{-1em}

\section{Conclusion}
We proposed a reuse of normalized initial classifier weights to mitigate the effect of catastrophic forgetting in memoryless IL.
A preliminary analysis showed that initial classifiers can be reused in latter incremental states, but a normalization step is needed to make them comparable.
We introduced a normalization component based on standardization which ensure fairness between classes learned in different IL states which are used together.
%Our method compares favorably to a standard handling of catastrophic forgetting via knowledge distillation ~\cite{DBLP:conf/eccv/LiH16,DBLP:conf/cvpr/RebuffiKSL17} and to a more sophisticated version which exploits distillation, inter-class separation and cosine normalization~\cite{DBLP:conf/cvpr/HouPLWL19}.
Our method compares favorably to a standard handling of catastrophic forgetting in~\cite{DBLP:conf/cvpr/HouPLWL19,DBLP:conf/eccv/LiH16,DBLP:conf/cvpr/RebuffiKSL17}.
Interestingly, our results indicate that distillation is only useful for small datasets and has a negative effect on larger datasets.
In this latter case, the use of simpler vanilla fine tuning backbone is more appropriate.
Note that the proposed method also improves the results obtained with classical distillation~\cite{DBLP:conf/eccv/LiH16,DBLP:conf/cvpr/RebuffiKSL17}.
However, the gap between classical learning remains important, and further efforts are needed towards reducing it.
The code is publicly available to facilitate reproducibility\footnote{\url{https://github.com/EdenBelouadah/class-incremental-learning/}}.

\noindent
\textbf{Acknowledgements}. This publication was made possible by the use of the FactoryIA supercomputer, financially supported by the Ile-de-France Regional Council.

\bibliography{main}

\newpage
\setcounter{page}{1}

\section*{Supplementary material}

\section*{1~~~Introduction}
In this supplementary material we provide: 
\begin{itemize}
    \item details about the evaluation datasets,
    \item implementation details for the tested methods,
    \item results with other normalization approaches,
    \item Error analysis for $FT$, $inFT^{mc}_{siw}$ and $LUCIR$.

\end{itemize}

\section*{2~~~Dataset details}
\label{sub:metho}
Four datasets that were designed for object, face, and landmark recognition are used here. 
The choice of significantly different tasks is essential to study the adaptability and robustness of the tested methods.
The main dataset statistics are provided in Table~\ref{tab_supp:dataset}.

\begin{itemize}
\item \textit{ILSVRC} ~\cite{DBLP:journals/ijcv/RussakovskyDSKS15} is a subset of 1000 $ImageNet$ classes used in the $ImagenetLSVRC$ challenges. It is constituted of leaves of the $ImageNet$ hierarchy which most often depict specific visual concepts.

\item \textit{VGGFace2}~\cite{DBLP:conf/fgr/CaoSXPZ18} is designed for face recognition. We selected 1000 classes having the largest number of associated images. Face cropping is done with MTCNN~\cite{DBLP:journals/spl/ZhangZLQ16} before further processing. 

\item \textit{Google~Landmarks}~\cite{DBLP:conf/iccv/NohASWH17} ($Landmarks$ below) is built for landmark recognition, and we selected 1000 classes having the largest number of associated images.

\item \textit{CIFAR100}~\cite{Krizhevsky09learningmultiple} is designed for object recognition and includes 100 basic level classes~\cite{ROSCH1976382}. 
\end{itemize}

\begin{table}[h!]
    \begin{center}
    \resizebox{0.47\textwidth}{!}
    {
    \begin{tabular}{|c|c|c|c|c|}
        \hline
         Dataset & train & test & $\mu$(train) & $\sigma$(train) \\ \hline
         ILSVRC     &  1,231,167  & 50,000  & 1231.16 & 70.18 \\ \hline 
         VGGFace2   &   491,746  & 50,000  & 491.74 & 49.37 \\ \hline 
         Landmarks  &  374,367  & 20,000 & 374.36 & 103.82 \\ \hline 
         CIFAR100   &  50,000  & 10,000  & 500.00 & 0.00 \\ \hline 

    \end{tabular}
    }
    \end{center}
    \caption{Main statistics for the evaluation datasets, $\mu$ is the mean number of images per class; and $\sigma$ is the standard deviation of the distribution of the number of images per class.}
    \label{tab_supp:dataset}
\end{table}

\section*{3~~~Implementation details}
A ResNet-18 architecture~\cite{DBLP:conf/cvpr/HeZRS16} with an SGD optimizer is used as a backbone for all the methods. $LUCIR$~\cite{DBLP:conf/cvpr/HouPLWL19} is run using the optimal parameters of the public implementation provided in the original paper. $LwF$~\cite{DBLP:conf/eccv/LiH16} is run using the code from~\cite{DBLP:conf/cvpr/RebuffiKSL17}.

$FT$ and its derivatives are based on the same fine-tuning backbone and are implemented in Pytorch~\cite{paszke2017automatic}. 
Training images are processed using randomly resized $224\times224$ crops, horizontal flipping, and are normalized afterward.
Given the difference in scale and the number of images between CIFAR100 and the other datasets, we found that a different parametrization was needed for this dataset. 
Note that the parameters' values presented below are largely inspired by the original ones given in~\cite{DBLP:conf/cvpr/HeZRS16}.

For CIFAR100, the first non-incremental state and $Full$ are run for 300 epochs with $batch~size=128$, $momentum=0.9$ and $weight~decay=0.0005$. 
The $lr$ is set to 0.1 and is divided by 10 when the error plateaus for 60 consecutive epochs. 
The incremental states of $FT$ are trained for 70 epochs with $batch~size=128$, $momentum=0.9$ and $weight~decay=0.0005$. 
The learning rate is set to $lr=0.1 / t$ at the beginning of each incremental state $\mathcal{Z}_t$ and is divided by 10 when the error plateaus for 15 consecutive epochs.
 
For ILSVRC, VGGFace2 and Landmarks, the first non-incremental state and $Full$ are run for 120 epochs with $batch~size=256$, $momentum=0.9$ and $weight~decay=0.0001$. The $lr$ is set to 0.1 and is divided by 10 when the error plateaus for 10 consecutive epochs. 
The incremental states of $FT$ are trained for 35 epochs with $batch~size=256$, $momentum=0.9$ and $weight~decay=0.0001$. 
The learning rate is set to $lr=0.1 / t$ at the beginning of each incremental state $\mathcal{Z}_t$ and is divided by 10 when the error plateaus for 5 consecutive epochs.

\section*{4~~~Results with other calibration methods}
Table~\ref{tab_supp:no_memory_results} provides results with mean and min-max normalization of weights in addition to $L2$ and $siw$. These two supplementary normalization techniques are defined below.

\begin{itemize}
    \item \textit{min-max normalization} - each dimension of the classifier is calculated using:
    
    \begin{equation}
    s_{k} = \frac{w_k-min(W)}{max(W)-min(W)}
    \end{equation}

    \item \textit{mean normalization} - each dimension of the classifier is calculated using
    
    \begin{equation}
    s_{k} = \frac{w_k-\mu(W)}{max(W)-min(W)}
    \end{equation}

\end{itemize}

Standardization provides the best performance for all tested configurations. 
Mean calibration is second best and has better performance compared to the L2-normalization already used in~\cite{scail2020}. Calibration with min-max is not effective and did not provide any good results.

\begin{table*}[t]
\begin{center}
\resizebox{\textwidth}{!}{
\begin{tabular}{|c|ccc|ccc|ccc|ccc||c|}
\hline
Dataset & \multicolumn{3}{|c|}{ILSVRC} & \multicolumn{3}{c|}{VGGFace2} & \multicolumn{3}{c|}{Landmarks}& \multicolumn{3}{c|}{CIFAR100} & \multirow{2}{*}{$G_{IL}$}  \\
\cline{1-13}
States & T=10  & T=20  & T=50  & T=10  & T=20 & T=50 & T=10  & T=20 & T=50 & T=10  & T=20 & T=50  & \\
\hline
{\small $inFT_{min-max}$  } & 3.3 & 10.0 & 7.1 & 4.7 & 20.1 & 18.5 & 17.2 & 12.2 & 6.3 & 19.9 & 18.3 & 20.7  & -55.52\\
{\small $inFT_{mean}$  } & 54.1 & 49.4 & 38.0 & 69.7 & 78.4 & 58.6 & 72.8 & 61.1 & 41.3 & 52.9 & 38.1 & 21.0 & -23.76\\
{\small $inFT_{L2}$ } & 51.6 & 43.3 & 34.5 & 76.8 & 66.8 & 55.1 & 61.4 & 52.5 & 39.2  & 47.5 & 39.3 & 22.5 & -26.80\\
{\small $inFT_{siw}$  } & \textbf{61.6} & \textbf{51.9} & \textbf{39.9} & \textbf{84.0} & \textbf{80.6} & \textbf{61.9} & \textbf{75.1} & \textbf{62.6} & \textbf{43.2} & \textbf{56.0} & \textbf{41.8} & \textbf{22.5}  & \textbf{-20.97}\\

\hline
$Full$ & 
\multicolumn{3}{|c|}{92.3}&
\multicolumn{3}{c|}{99.2} & \multicolumn{3}{c|}{99.1} & \multicolumn{3}{c|}{91.2}  & 
- \\
\hline
\end{tabular}
}
\end{center}

%\vspace{-1em}
    \caption{Top-5 average IL accuracy (\%) for the min-max and mean normalization tested in addition to L2 and standardization, with T=\{10, 20, 50\} incremental states. Best results are in bold.
    }
\label{tab_supp:no_memory_results}
\end{table*}

\section*{5~~~Error analysis}
Following~\cite{scail2020}, in Table~\ref{tab_supp:errors}, we provide top-1 correct and wrong classifications for: (1) $FT$ - the simplest method tested, (2) $LUCIR$ - the best existing method (3) $inFT^{mc}_{siw}$ - the proposed method. 
The analysis is done for the large dataset ILSVRC, with $T=20$ states. 
$c(p)$ and $c(n)$ are the correct classification for past/new classes. 
$e(p,p)$ and $e(p,n)$ are erroneous classifications for test samples of past classes mistaken for other past classes and new classes respectively. 
$e(n,p)$ and $e(n,n)$ are erroneous classifications for test samples of new classes mistaken for past classes and other new classes respectively. 
Note that the percentages on the first three and last three lines of each table sum up to 100\%.
Since the number of test images varies across IL states, percentages are calculated separately for test images of past and new classes in each $\mathcal{Z}_t$ to get a quick view of the relative importance of each type of errors. $c(p)$, $e(p,p)$, and $e(p,n)$ sum to 100\% on each column, as do $c(n)$, $e(n,n)$, and $e(n,p)$. The analysis shows that vanilla $FT$ suffers from a total forgetting of the past classes since all their test images are wrongly classified. 
The effect of catastrophic forgetting is obvious in the way that 100\% of past classes are mistakenly classified as belonging to new classes. 
Equally important, standardization of the initial weights not only reduces forgetting, but also reduces considerably the confusions among new classes.
The comparison of $LUCIR$ and $inFT^{mc}_{siw}$ shows that the first method is better at classifying test samples of new classes but has worse behavior for test samples of past classes.
$LUCIR$ $c(p)$ scores are better for the first three iterations but fall behind those of $inFT^{mc}_{siw}$ afterwards.
Note that both methods are strongly affected by catastrophic forgetting toward the end of the incremental process, with top-1 accuracy at 6\% and 11.8\% for $LUCIR$ and $inFT^{mc}_{siw}$ respectively. 
This finding indicates that, while both distillation in $LUCIR$ and classifier weights replay $inFT^{mc}_{siw}$ have a slight positive effect, memoryless IL remains a very challenging task.
It is also interesting that the distribution of errors is different. 
$LUCIR$ fails to ensure fairness between past and new classes since $e(p,n)$ are much more frequent than $e(p,p)$. 
$inFT^{mc}_{siw}$ is less biased toward new classes but produces a large number of confusions between past classes ($e(p,p)$.

\begin{table}[t]\centering
\resizebox{\textwidth}{!}{
    \begin{tabular}{c|c|c@{\hskip 0.12in}c@{\hskip 0.12in}c@{\hskip 0.12in}c@{\hskip 0.12in}c@{\hskip 0.12in}c@{\hskip 0.12in}c@{\hskip 0.12in}c@{\hskip 0.12in}c@{\hskip 0.12in}c@{\hskip 0.12in}c@{\hskip 0.12in}c@{\hskip 0.12in}c@{\hskip 0.12in}c@{\hskip 0.12in}c@{\hskip 0.12in}c@{\hskip 0.12in}c@{\hskip 0.12in}c@{\hskip 0.12in}c}
  %       & & \multicolumn{19}{c}{\textbf{Incremental states}}\\
  \hline
         \multicolumn{2}{c}{\textbf{Incremental states}}& $\mathcal{Z}_1$ & $\mathcal{Z}_2$  & $\mathcal{Z}_3$ & $\mathcal{Z}_4$ & $\mathcal{Z}_5$ & $\mathcal{Z}_6$ & $\mathcal{Z}_7$ & $\mathcal{Z}_8$ & $\mathcal{Z}_9$  & $\mathcal{Z}_{10}$ & $\mathcal{Z}_{11}$ & $\mathcal{Z}_{12}$ & $\mathcal{Z}_{13}$ & $\mathcal{Z}_{14}$ & $\mathcal{Z}_{15}$ & $\mathcal{Z}_{16}$ & $\mathcal{Z}_{17}$ & $\mathcal{Z}_{18}$ & $\mathcal{Z}_{19}$  \\
        \midrule
        %\multicolumn{21}{c}{ILSVRC}\\
        %\midrule
\parbox[t]{2mm}{\multirow{6}{*}{\rotatebox[origin=c]{90}{$FT$}}}   

&$c(p)$  & 0.0 & 0.0 & 0.0 & 0.0 & 0.0 & 0.0 & 0.0 & 0.0 & 0.0 & 0.0 & 0.0 & 0.0 & 0.0 & 0.0 & 0.0 & 0.0 & 0.0 & 0.0 & 0.0 \\
&$e(p, p)$  & 0.0 & 0.0 & 0.0 & 0.0 & 0.0 & 0.0 & 0.0 & 0.0 & 0.0 & 0.0 & 0.0 & 0.0 & 0.0 & 0.0 & 0.0 & 0.0 & 0.0 & 0.0 & 0.0 \\
&$e(p, n)$  & 100.0 & 100.0 & 100.0 & 100.0 & 100.0 & 100.0 & 100.0 & 100.0 & 100.0 & 100.0 & 100.0 & 100.0 & 100.0 & 100.0 & 100.0 & 100.0 & 100.0 & 100.0 & 100.0 \\
&$c(n)$  & 87.8 & 87.28 & 90.48 & 91.4 & 90.44 & 87.92 & 89.64 & 88.12 & 87.24 & 89.68 & 89.72 & 90.16 & 90.6 & 89.8 & 87.84 & 92.4 & 89.56 & 89.28 & 87.52 \\
&$e(n, n)$  & 12.2 & 12.72 & 9.52 & 8.6 & 9.56 & 12.08 & 10.36 & 11.88 & 12.76 & 10.32 & 10.28 & 9.84 & 9.4 & 10.2 & 12.16 & 7.6 & 10.44 & 10.72 & 12.48 \\
&$e(n, p)$  & 0.0 & 0.0 & 0.0 & 0.0 & 0.0 & 0.0 & 0.0 & 0.0 & 0.0 & 0.0 & 0.0 & 0.0 & 0.0 & 0.0 & 0.0 & 0.0 & 0.0 & 0.0 & 0.0 \\

\midrule
\parbox[t]{2mm}{\multirow{6}{*}{\rotatebox[origin=c]{90}{$inFT_{siw}^{mc}$}}}  

&$c(p)$  & 38.4 & 27.0 & 33.2 & 31.3 & 29.0 & 22.0 & 20.1 & 15.0 & 17.9 & 14.7 & 17.7 & 16.5 & 15.3 & 13.1 & 13.2 & 14.0 & 14.1 & 12.5 & 11.8 \\
&$e(p, p)$  & 22.7 & 15.0 & 41.4 & 41.9 & 60.7 & 48.5 & 51.8 & 31.9 & 60.2 & 40.7 & 68.1 & 62.6 & 66.8 & 48.2 & 47.2 & 66.9 & 64.9 & 52.7 & 50.0 \\ 
&$e(p, n)$  & 38.9 & 58.0 & 25.4 & 26.8 & 10.3 & 29.5 & 28.1 & 53.0 & 21.9 & 44.6 & 14.1 & 20.9 & 17.9 & 38.7 & 39.6 & 19.1 & 21.0 & 34.8 & 38.2 \\
&$c(n)$  & 75.8 & 82.7 & 75.7 & 75.8 & 67.2 & 75.1 & 77.4 & 83.8 & 69.8 & 83.2 & 68.6 & 76.1 & 70.5 & 82.0 & 78.4 & 76.2 & 72.6 & 80.4 & 80.3 \\
&$e(n, n)$  & 8.5 & 11.5 & 4.2 & 3.1 & 1.8 & 6.8 & 4.6 & 9.8 & 4.5 & 7.9 & 3.2 & 3.5 & 3.1 & 6.5 & 8.3 & 2.7 & 3.8 & 5.4 & 7.7 \\
&$e(n, p)$  & 15.7 & 5.8 & 20.0 & 21.1 & 31.0 & 18.0 & 18.0 & 6.4 & 25.7 & 8.9 & 28.2 & 20.4 & 26.4 & 11.5 & 13.3 & 21.1 & 23.6 & 14.1 & 12.0\\

%\midrule
%\parbox[t]{2mm}{\multirow{6}{*}{\rotatebox[origin=c]{90}{$LwF$}}}

%&$c(p)$  & - & - & - & - & - & - & - & - & - & - & - & - & - & - & - & -  & - & - & -\\
%&$e(p, p)$  & - & - & - & - & - & - & - & - & - & - & - & - & - & - & - & - & - & - & -\\
%&$e(p, n)$  & - & - & - & - & - & - & - & - & - & - & - & - & - & - & -  & - & - & - & -\\
%&$c(n)$  & - & - & - & - & - & - & - & - & - & - & - & - & - & - & - & -  & - & - & -\\
%&$e(n, n)$ & - & - & - & - & - & - & - & - & - & - & - & - & - & - & - &  - & - & - & -\\
%&$e(n, p)$  & - & - & - & - & - & - & - & - & - & - & - & - & - & - &  - & - & - & - & -\\

\midrule
\parbox[t]{2mm}{\multirow{6}{*}{\rotatebox[origin=c]{90}{$LUCIR$}}}

&$c(p)$  & 66.1 & 46.9 & 33.5 & 26.7 & 23.2 & 19.0 & 15.1 & 13.3 & 11.8 & 9.9 & 9.1 & 8.3 & 7.9 & 8.0 & 7.5 & 6.7 & 6.5 & 6.0 & 6.0 \\
&$e(p, p)$  & 4.2 & 10.1 & 14.7 & 20.4 & 26.6 & 25.8 & 24.1 & 27.5 & 27.9 & 28.1 & 28.3 & 28.8 & 29.8 & 31.5 & 29.3 & 30.8 & 29.6 & 29.6 & 30.6 \\ 
&$e(p, n)$  & 29.8 & 42.9 & 51.8 & 52.9 & 50.2 & 55.3 & 60.8 & 59.2 & 60.3 & 62.0 & 62.6 & 63.0 & 62.3 & 60.5 & 63.2 & 62.5 & 63.9 & 64.4 & 63.4 \\
&$c(n)$  & 78.3 & 79.7 & 82.2 & 82.2 & 82.4 & 78.2 & 82.6 & 81.5 & 79.0 & 84.5 & 82.7 & 83.4 & 84.1 & 82.9 & 81.2 & 86.2 & 82.8 & 83.3 & 81.2 \\
&$e(n, n)$  & 16.0 & 15.5 & 13.5 & 11.4 & 12.2 & 15.2 & 12.3 & 13.0 & 14.5 & 11.4 & 11.9 & 11.9 & 11.2 & 11.5 & 14.2 & 9.0 & 11.7 & 12.1 & 13.8 \\
&$e(n, p)$  & 5.6 & 4.8 & 4.4 & 6.4 & 5.3 & 6.6 & 5.2 & 5.5 & 6.5 & 4.1 & 5.4 & 4.8 & 4.6 & 5.6 & 4.6 & 4.8 & 5.5 & 4.6 & 5.0 \\

\hline     

\hline     
\end{tabular}
}

\vspace{1em}

\caption{Top-1 correct and wrong classification for $FT$, $inFT^{mc}_{siw}$ and $LUCIR$ for ILSVRC with $T=20$. 
    }
\label{tab_supp:errors}
\end{table}

\clearpage 

\bibliography{main2}

\end{document}